\documentclass{article}
\usepackage{spconf,amsmath,graphicx}
\usepackage{hyperref} 
\usepackage{enumitem}
\setlist{nosep, leftmargin=14pt}
\usepackage{cleveref}
\usepackage{tabu}                      % only used for the table example
\usepackage{booktabs}                  % only used for the table example
\usepackage{lipsum}                    % used to generate placeholder text
\usepackage{mwe}                       % used to generate placeholder figures
\usepackage{listings}
\usepackage{caption}

\usepackage{color}

\usepackage{mwe} % to get dummy images

\title{Designing a Convolutional Neural Network for High-Accuracy Oral Cavity Squamous Cell Carcinoma (OCSCC) Detection}

\name{Vishal Manikanden$^{1}$ \qquad Aniketh Bandlamudi $^{1}$ \qquad Daniel Haehn$^{1}$}

\address{$^{1}$ University of Massachusetts Boston, 100 Morrissey Blvd, Boston, MA 02125}

\begin{document}
%\ninept
%
\maketitle
\begin{abstract}
Oral Cavity Squamous Cell Carcinoma (OCSCC) is the most common type of head and neck cancer. Due to the subtle nature of its early stages, deep and hidden areas of development, and slow growth, OCSCC often goes undetected, leading to preventable deaths. However, properly trained Convolutional Neural Networks (CNNs), with their precise image segmentation techniques and ability to apply kernel matrices to modify the RGB values of images for accurate image pattern recognition, would be an effective means for early detection of OCSCC. Pairing this neural network with image capturing and processing hardware would allow increased efficacy in OCSCC detection. The aim of our project is to develop a Convolutional Neural Network trained to recognize OCSCC, as well as to design a physical hardware system to capture and process detailed images, in order to determine the image quality required for accurate predictions. A CNN was trained on 4293 training images consisting of benign and malignant tumors, as well as negative samples, and was evaluated for its precision, recall, and Mean Average Precision (mAP) in its predictions of OCSCC. A testing dataset of randomly assorted images of cancerous, non-cancerous, and negative images was chosen, and each image was altered to represent 5 common resolutions. This test data set was thoroughly analyzed by the CNN and predictions were scored on the basis of accuracy. The designed enhancement hardware was used to capture detailed images, and its impact was scored. An application was developed to facilitate the testing process and bring open access to the CNN. Images of increasing resolution resulted in higher-accuracy predictions on a logarithmic scale, demonstrating the diminishing returns of higher pixel counts.
\end{abstract}
\section{Introduction}
\label{sec:intro}
Oral Cavity Squamous Cell Carcinoma (OCSCC) has become an increasingly significant global health concern, with high mortality rates due to late diagnosis. This type of cancer is the most common type of head and neck cancer, and is considered extremely aggressive because of its metastatic properties. Traditional, commonly used diagnostic methods rely on manual screening done by health professionals. These methods of diagnosis are known to be expensive, and time consuming. Furthermore, they remain inaccessible to the vast majority of people, especially those residing in rural or under-resourced communities \cite{imbesi2023oral}. 

Computer Vision (CV) poses an extremely effective means of early OCSCC detection. Specifically, Convolutional Neural Networks (CNNs) are a subtype of CV that focus on deep neural networks for image segmentation. These neural networks mimic the method that the human brain employs to process visual information \cite{o2015introduction}. Each CNN has various convolutional layers that each serve a certain functionality in developing the overall CV model. The first layer is known as the input layer, where all images are converted to a matrix of RGB pixels. Following this conversion, the image enters the convolution layer, in which a kernel image filter is convolved over the image to manipulate the image. This kernel is a matrix of numerical values that detects image features and extracts them by computing the dot product of each element of its matrix with the corresponding element of the image pixel matrix. The neural network then passes the image to the activation function, which applies a function to each individual neuron (unit) of the neural network. This function is meant to prevent the equivalent of a linear regression of the model by introducing nonlinearity into the model. This is achieved by modifying the weight of each neuron until a weighted sum of all neurons is achieved for each individual image. The Rectified Linear Unit (ReLU) function (Fig. 1) is an extremely effective activation function for CNNs because of its efficient property and ability to mitigate the vanishing gradient problems of other activation functions, in which the image is saturated in the positive area, which reduces image quality and detail. Following this layer, the pooling layer applies a pooling function to reduce the dimensions of image matrices to increase the efficiency of image processing. In a CNN, this requirement is fulfilled using the max pooling function, in which the neural network determines the maximum pixel value of each region of a specified size \cite{yamashita2018convolutional}. This is followed by the fully connected layers, in which the neural network is finalized by connecting each neuron in the previous layer, leading to a dense array of neuron-connected layers. Finally, the output layer produces the prediction of the CNN. In the application of OCSCC detection, this output would be a value representing the state of the mouth as either cancerous, non-cancerous, or negative, as well as a percentage of certainty of the model in its prediction \cite{pianykh2020continuous}.

The inputs to the CNN must be as accurate and detailed as possible in order for the kernel of the neural network to properly extract the features of the image. Therefore, hardware is required to assist the user in the image-capturing process. This hardware must be able to conform to the camera module that is being used to capture the image of the mouth. It would create a tight fit in the user’s mouth such that the image that is captured properly represents the user’s mouth. By gaining more accurate images, the neural network would better be able to assign weights to individual neurons based on the patterns in the image matrix that are present, thereby increasing the efficacy of the predictions of the model \cite{mascitti2018overview}.

This research project was aimed to determine the impact of image resolution on the efficacy of convolutional neural networks in the detection of OCSCC when inputting an image of an oral cavity taken with and without image-capturing hardware. The independent variable was the resolution of the images that were inputted to the model, as well as the use of the image-capturing hardware, while the dependent variable was the accuracy of the model in its predictions of OCSCC. Our hypothesis states that increasing the resolution of input images in the computer vision model will lead to higher predictive accuracy by providing more detailed and informative visual data; furthermore, incorporating the image-capturing hardware would enhance accuracy by improving data capture quality.

\section{MATERIALS \& METHODS}
\label{sec:related_work}

\subsection{Data Collection}
A dataset of 3275 images was collected from the Mendeley Dataset, which included images of benign and malignant tumors. All cancerous and non-cancerous images were identified and isolated. In addition, a general image dataset of 1018 images was collected from Google Open Images, which contained general images that were not of an oral cavity and represented the negative tag.

\subsection{Convolutional Neural Network Design:}

A deep convolutional neural network was developed to accurately determine the cancerous or non-cancerous state of a mouth image. First, the dataset of images was compiled, and all images were classified into the labels of cancerous, non-cancerous, or negative. The training layers of the neural network were determined and set, and the initial weights and neurons were determined. In the input layer, each individual pixel of the training image was given a neuron. The convolution layer contained the kernels that would manipulate each neuron to emphasize the defining features of the image that would either determine the diagnosis or define the image as a negative image. The ReLU activation function was defined as the next layer, as this back propagation technique was fitting for the purpose as an image classification model. The fully connected layers following this properly connected each neuron so that the model would be able to develop a weighted sum by calculating the summation of all weighted connected neurons. This sum was utilized in the final layer to output the model’s prediction based on the classification with the largest weight, as well as to determine its percentage of certainty with its prediction.

\subsection{Convolutional Neural Network Training}
In order to make the neural network as accurate and consistent as possible, we found the optimal ratio of batch size to epochs to iterations that would be used to train the model. The batch size determined the number of images that would be processed before the weights of the neurons of the model were updated. The number of epochs represented the number of passes the neural network would make over the training data. The number of iterations per epoch was determined by dividing the number of training images (4293) by the batch size. An optimal ratio of these three values was crucial for making the CNN increasingly accurate in order to spend an amount of time that would allow the weights to become accurate without the model overcorrecting. The optimal ratio that we calculated to achieve this was 32:20:134, which meant that there would be around 2700 changes to the weights of all neurons combined. Following this, 8 hours were allocated to the training of the model, which was precisely conducted based on this ratio, with the ReLU function being applied to properly update the weights of each neuron.

\subsection{Convolutional Neural Network Evaluation}
In order to evaluate the neural network, 3 main metrics were utilized. These metrics were determined based on the performance of the CNN during the rigorous evaluation phase of the training. The precision metric represented the percentage of the CNN’s predictions that were correct. The recall metric represented the percentage of test cases that were properly predicted. The Mean Average Precision (mAP) metric represented the overall accuracy of the model across all the different test cases. A testing dataset of 50 images containing cancerous, non-cancerous, and negative images was randomly chosen. For each image, the Pixelmator Pro software was utilized to adjust the resolution of each image to 144p, 360p, 720p, 1080p, and 1440p, totaling a testing dataset of 250 images. Each adjusted image was used to evaluate the efficacy of the CNN. In order to properly evaluate the model, an iOS application was developed using the Swift programming language and the XCode development environment. This application provided a simple means of accessing the CNN and inputting images to obtain its predictions.

\begin{figure}[h!]
 \centering
\includegraphics[width=0.35\columnwidth]{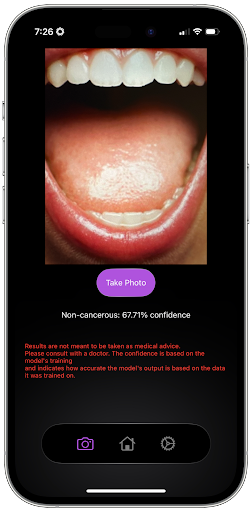} \caption{Apple iOS Application displaying an example (non-cancerous) scan with its accompanying result and model confidence.}
 \label{fig:boostlets_sam}
\end{figure}

\subsection{Hardware Development}
Autodesk Fusion 360, a computer-aided design (CAD) software, was utilized to develop a model and rendering of the hardware that would assist in the image capturing process. This hardware contained a mouth-fitting piece that would be able to fit inside a user’s mouth and provide the camera with a clear depiction of the mouth. 
\begin{figure}[h!]
 \centering
\includegraphics[width=0.5\columnwidth]{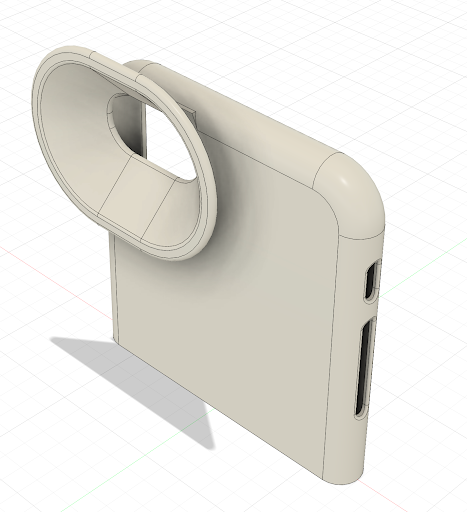} \caption{Rendering of the image enhancing hardware attachment used to aid with the scanning of the oral cavity.}
 \label{fig:boostlets_sam}
\end{figure}
The hardware was designed to adapt and fit around the camera module of an iPhone SE, similar to the form-factor of a phone case.
\begin{figure}[h!]
 \centering
\includegraphics[width=0.5\columnwidth]{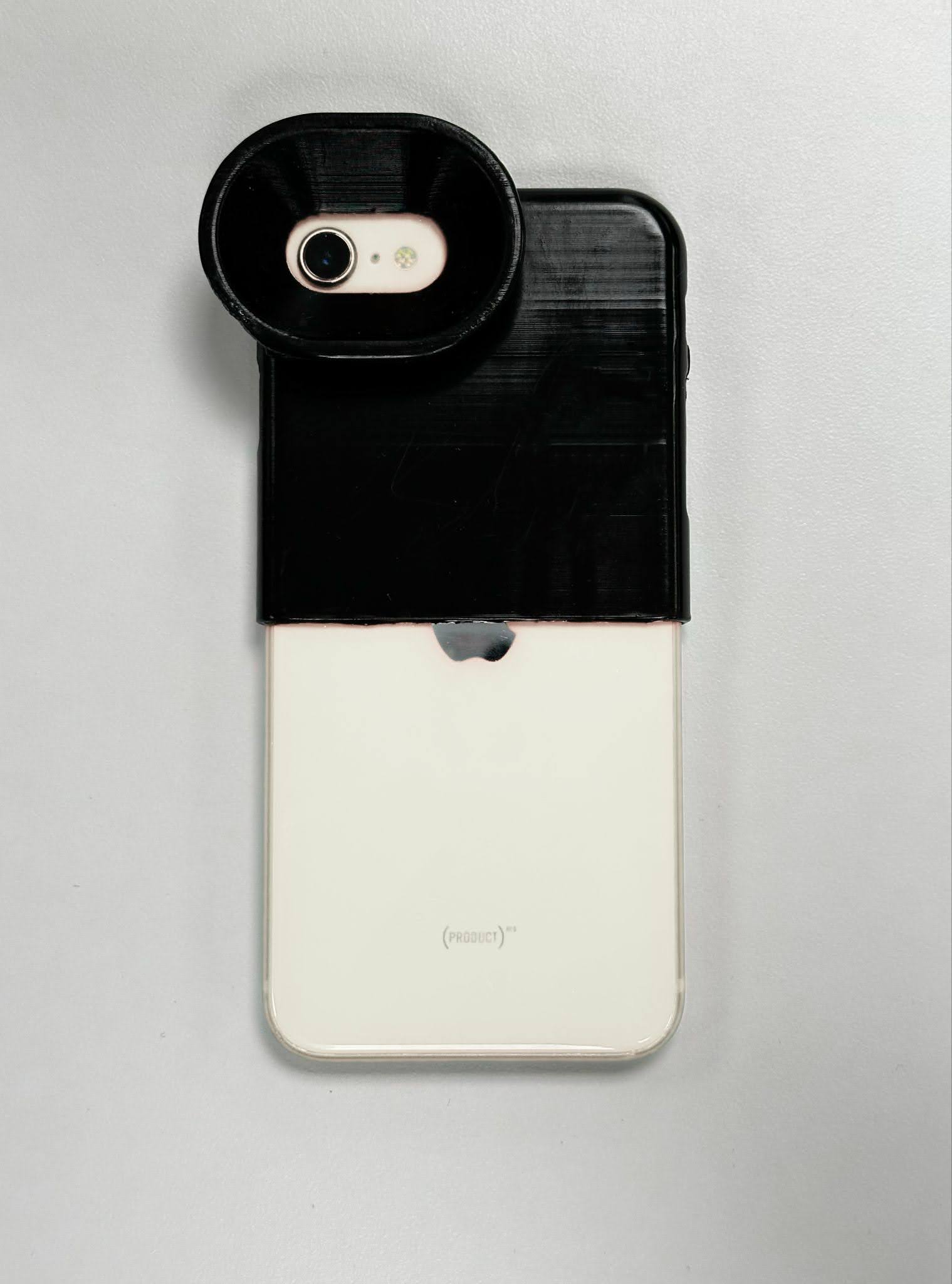} \caption{The fit of the hardware around an iPhone SE model, demonstrating its ease of attachment and removal.}
 \label{fig:boostlets_sam}
\end{figure}
An Ultimaker Cura 3D printer was utilized to 3D print the piece. 
\begin{figure}[h!]
 \centering
\includegraphics[width=0.5\columnwidth]{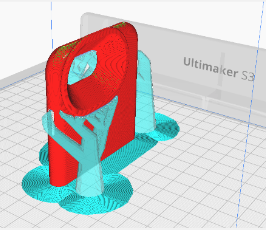} \caption{Processing of the model with the supports necessary to print in the Cura printer.}
 \label{fig:boostlets_sam}
\end{figure}
The impact of the hardware was tested by taking images of a non-cancerous mouth both with and without the piece, and modifying to the same 5 resolutions to be evaluated by the model. This would determine the utility of this image-capturing hardware by determining the difference in accuracy of the model with and without the hardware. 

\section{Results}
\label{sec:usage}
The results of the efficacy of the convolutional neural network are presented below. The evaluation of the 3 metrics used to test the CNN, as well as its accuracy percentage, are depicted. In addition, the efficacy of the image-capturing hardware is discussed. Finally, the effect of altering image resolutions is graphed and evaluated as an average across all classification types and use or lack of use of the hardware.

\subsection{Convolutional Neural Network Evaluation Metrics}
\label{sec:cnn_evaluation_metrics}
The precision, recall and average precision of the model is presented in Figure 5. These metrics were measured during the training of the model, where 20\% of the dataset was allocated as the evaluation dataset.
\begin{figure}[h!]
 \centering
\includegraphics[width=1\columnwidth]{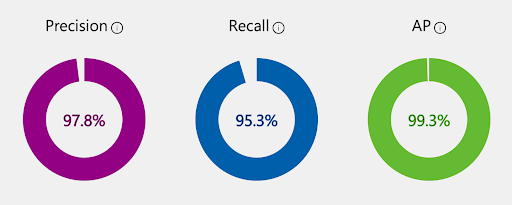} \caption{Precision signifies the proportion of correctly identified positive instances against all predicted positives. Recall measures the proportion of actual positive instances correctly detected by the model. Mean Average Precision (AP) represents the area under the precision-recall curve, which is a comprehensive assessment of the model’s performance across varying thresholds.}
 \label{fig:boostlets_sam}
\end{figure}

\subsection{Convolutional Neural Network Accuracy}
\label{sec:cnn_accuracy}
The accuracy of the model across the testing dataset of 50 images of cancerous, non-cancerous, and general images is depicted in Figure 6. The graph represents the accuracy of the model across all 5 resolutions.
\begin{figure}[h!]
 \centering
\includegraphics[width=1\columnwidth]{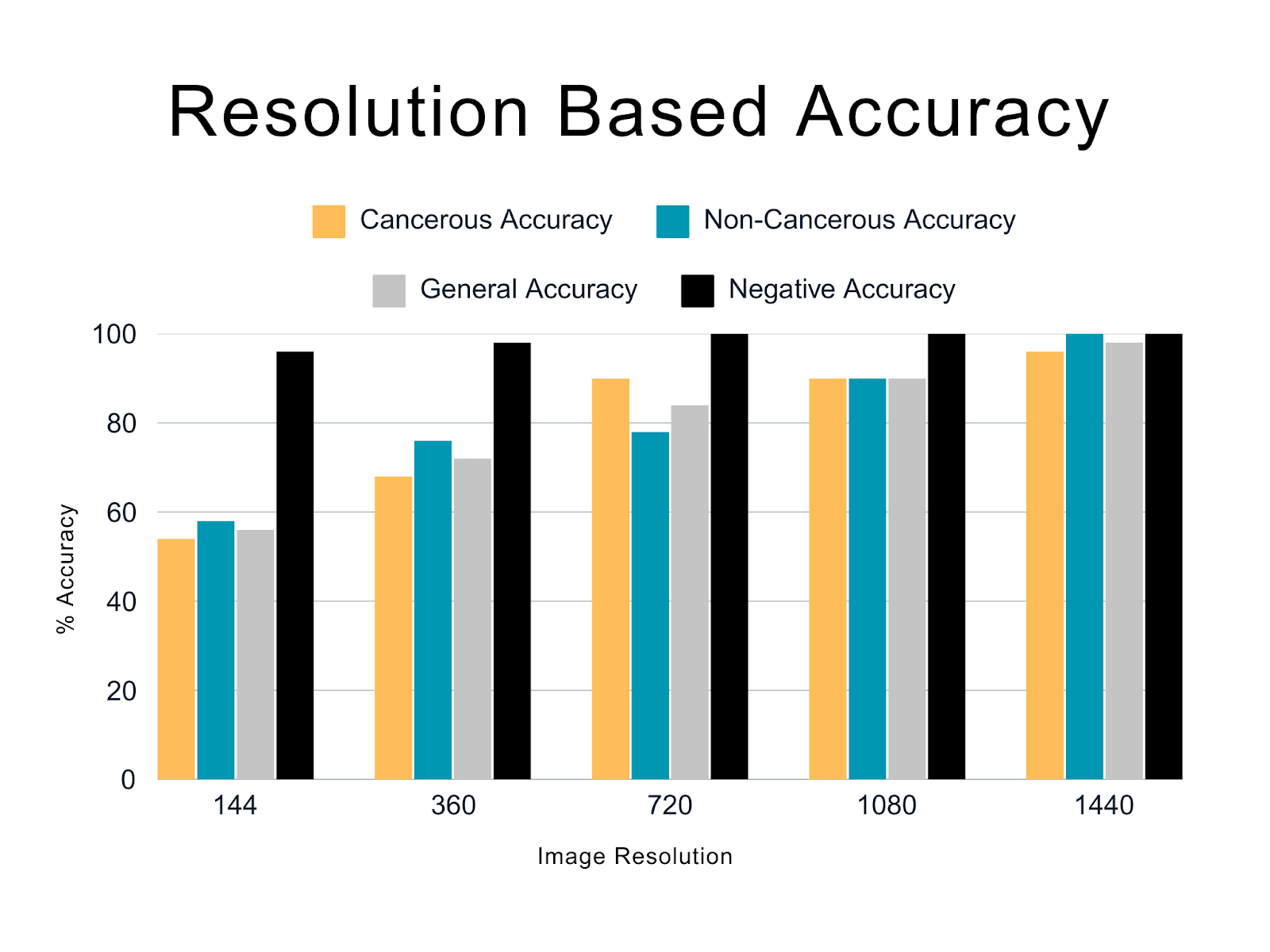} \caption{Displays the average accuracy of the CNN in predicting cancerous images, non cancerous images, and negative images across the 5 resolutions.}
 \label{fig:boostlets_sam}
\end{figure}

\subsection{Impact of Image-Capturing Hardware}
\label{sec:impact_image_capturing_hardware}
The effect of utilizing the image-capturing hardware across all 5 resolutions is displayed in Figure 7. This bar graph offers a comprehensive overview of the increased accuracy of the CNN with the image-capturing hardware.
\begin{figure}[h!]
 \centering
\includegraphics[width=1\columnwidth]{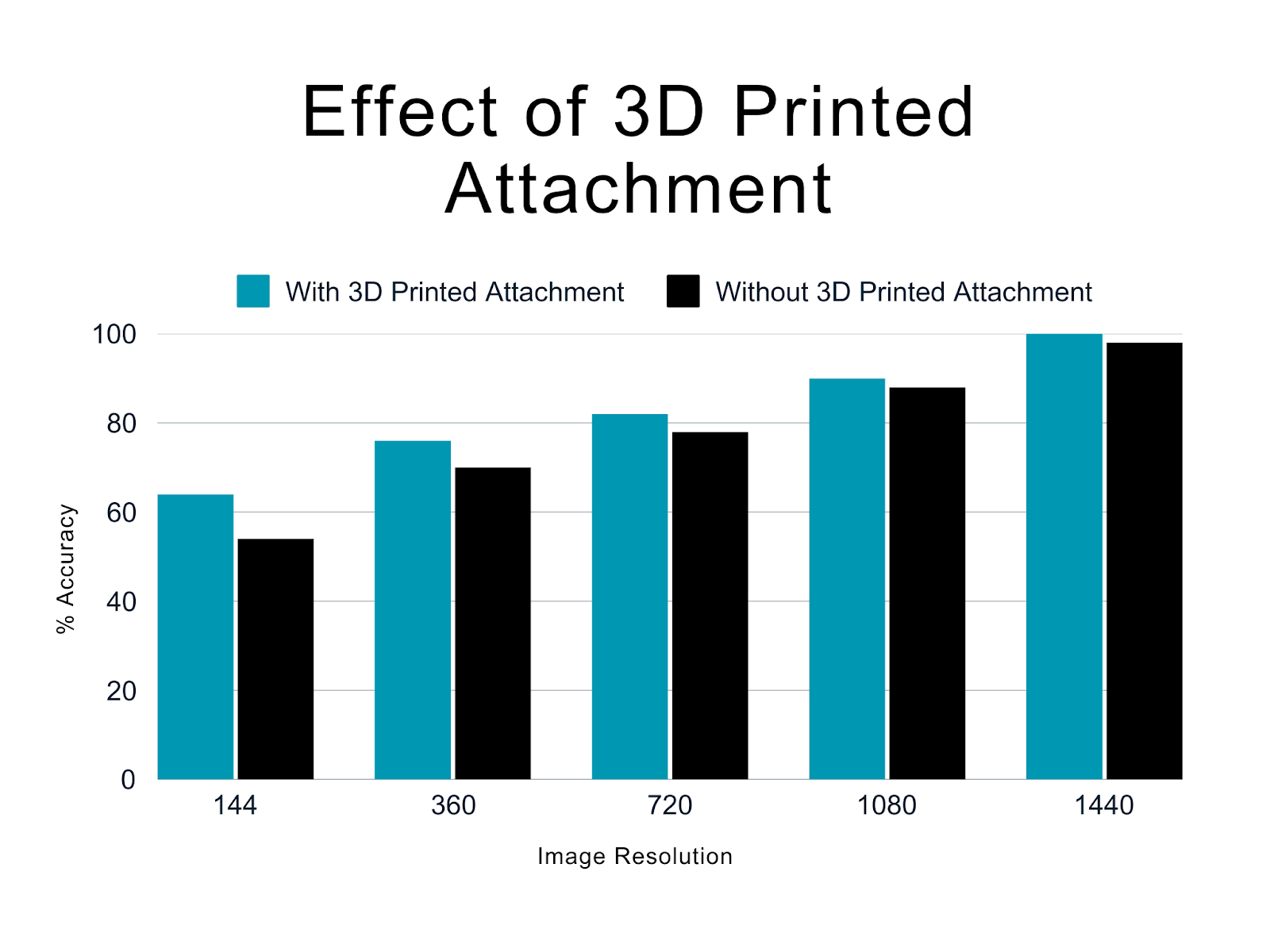} \caption{Displays the higher performance in average accuracy of images with the image-capturing hardware across the 5 resolutions.}
 \label{fig:boostlets_sam}
\end{figure}

\subsection{Overall Effect of Altering Image Resolution on Prediction Accuracy}
\label{sec:effect_image_resolution}
The effect of utilizing the image-capturing hardware across all 5 resolutions is displayed in Figure 8. This bar graph offers a comprehensive overview of the increased accuracy of the CNN with the image-capturing hardware.
\begin{figure}[h!]
 \centering
\includegraphics[width=1\columnwidth]{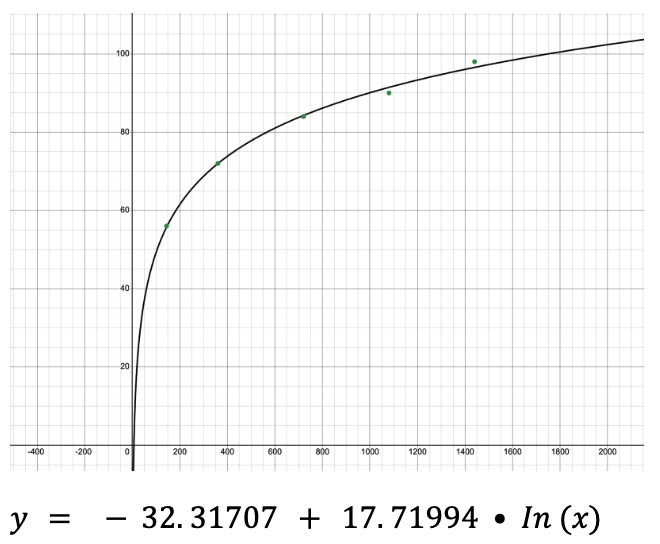} \caption{Displays the logarithmic trend in average accuracy in all three variations of images. The R² Coefficient of determination is 0.996, displaying the strength of the logarithmic model for this data. }
 \label{fig:boostlets_sam}
\end{figure}

\section{DISCUSSION \& CONCLUSIONS}

The purpose of this study was to determine the efficacy of a combination of a convolutional neural network and image-capturing hardware in the detection of Oral Cavity Squamous Cell Carcinoma. A convolutional neural network was developed using a comprehensive system of training layers and a dataset consisting of thousands of images representing cancerous and non-cancerous images of mouths, as well as negative images, and hardware was designed and 3D printed to assist in the image-capturing process. The resolution of testing images was altered and inputted to the model to evaluate its prediction accuracy as a function of image quality. Following this procedure, we concluded that overall, higher image resolution leads to increasing model accuracy for the detection of OCSCC, but the impacts were only present on a logarithmic scale. This means that the return for higher resolution images in resolution decreases as with higher resolutions. The detection of negative images was largely consistent across all resolutions. In addition, we determined that utilizing the image-capturing hardware increases the accuracy of the model significantly due to higher-quality images with more information being captured. The results met the expected outcomes, without significant bias. A limitation of this study lies in the dataset. While extensive, it may not fully represent the vast diversity of real world cases. This may lead to certain edge cases not being accurately detected. The logarithmic trend in accuracy improvement suggests that beyond a certain resolution threshold, there may not be significant gains in accuracy through further increases in image quality. This indicates the need for an optimal balance between resolution and computational efficiency.

\section{FUTURE IMPACT}

\subsection{Early Detection of OCSCC}
This study demonstrates the potential of CNNs and hardware for the detection of OCSCC. The efficacy of the model from simply phone-captured images underscores its massive potential in diagnosing OCSCC preemptively. In addition, the application that we designed provides ease of access for users to the model, with great potential to impact rural and underprivileged areas with this increased ability for OCSCC detection.

\subsection{Future Work}
We will continue to improve our CNN by introducing training layers of increasing efficacy and impact while employing comprehensive and larger datasets. We will continue to modify the image-processing hardware to allow users to capture images of increasing quality and detail by introducing a fully adaptable system and constructing the hardware out of machined aluminum. We will improve the user interface of the application, expanding it to the Android and web space to allow users of any device to easily access our CNN.

\bibliographystyle{IEEEbib}
\bibliography{strings,refs}

\end{document}